\documentclass[letterpaper, 10 pt, journal, twoside]{IEEEtran}

\makeatletter
\let\NAT@parse\undefined
\makeatother

\IEEEoverridecommandlockouts                              %

\usepackage[numbers,sort&compress]{natbib}

\usepackage{multirow}
\usepackage{multicol}
\usepackage[bookmarks=true]{hyperref}
\usepackage{xcolor}
\usepackage{tablefootnote} %
\usepackage{subcaption}
\usepackage{graphicx}
\usepackage{float}
\usepackage{amssymb}
\usepackage{amsmath}
\DeclareMathOperator*{\argmax}{argmax}
\DeclareMathOperator*{\argmin}{argmin}

\usepackage{sidecap}

\usepackage{tikz}
\def\checkmark{\tikz\fill[scale=0.4](0,.35) -- (.25,0) -- (1,.7) -- (.25,.15) -- cycle;} 

\begin{document}
\bstctlcite{BSTcontrol}

\title{
ViHOPE: Visuotactile In-Hand Object 6D Pose Estimation with Shape Completion
}

\author{Hongyu Li$^{1,2}$, Snehal Dikhale$^{1}$, Soshi Iba$^{1}$, and Nawid Jamali$^{1}$%

\thanks{Manuscript received: April 18, 2023; Revised July 19, 2023; Accepted August 25, 2023.}%
\thanks{This paper was recommended for publication by Editor Markus Vincze upon evaluation of the Associate Editor and Reviewers' comments.} %
\thanks{$^{1}$The authors are with Honda Research Institute USA, Inc. 
        {\tt\footnotesize \{snehalsubhash\_dikhale, siba, njamali\}@honda-ri.com}}%
\thanks{$^{2} $Hongyu Li is with Brown University {\tt\footnotesize hongyu@brown.edu}. This work was completed when he was an intern at Honda Research Institute USA, Inc.}%
\thanks{Digital Object Identifier (DOI): see top of this page.}
}

\markboth{IEEE Robotics and Automation Letters. Preprint Version. Accepted August, 2023}
{Li \MakeLowercase{\textit{et al.}}: ViHOPE} 

\maketitle

\newcommand{\lhy}[1]{\textcolor{red}{LHY: #1}}

\newcommand{\syn}[1]{\textcolor{blue}{Synthetic}}
\newcommand{\real}[1]{\textcolor{green}{Real World}}
\newcommand{\nj}[1]{\textcolor{magenta}{#1}}

\newcommand{\optional}[1]{\textcolor{magenta}{#1}}

\begin{abstract}
In this letter, we introduce ViHOPE, a novel framework for estimating the 6D pose of an in-hand object using visuotactile perception. Our key insight is that the accuracy of the 6D object pose estimate can be improved by explicitly completing the shape of the object. 
To this end, we introduce a novel visuotactile shape completion module that uses a conditional Generative Adversarial Network to complete the shape of an in-hand object based on volumetric representation. This approach improves over prior works that directly regress visuotactile observations to a 6D pose. By explicitly completing the shape of the in-hand object and jointly optimizing the shape completion and pose estimation tasks, we improve the accuracy of the 6D object pose estimate.
We train and test our model on a synthetic dataset and compare it with the state-of-the-art. 
In the visuotactile shape completion task, we outperform the state-of-the-art by 265\% using the Intersection of Union metric and achieve 88\% lower Chamfer Distance. In the visuotactile pose estimation task, we present results that suggest our framework reduces position and angular errors by 35\% and 64\%, respectively. Furthermore, we ablate our framework to confirm the gain on the 6D object pose estimate from explicitly completing the shape. Ultimately, we show that our framework produces models that are robust to sim-to-real transfer on a real-world robot platform. 
\end{abstract}

\section{Introduction}

\begin{figure}[ht!]
    \centering
    \includegraphics[width=\columnwidth]{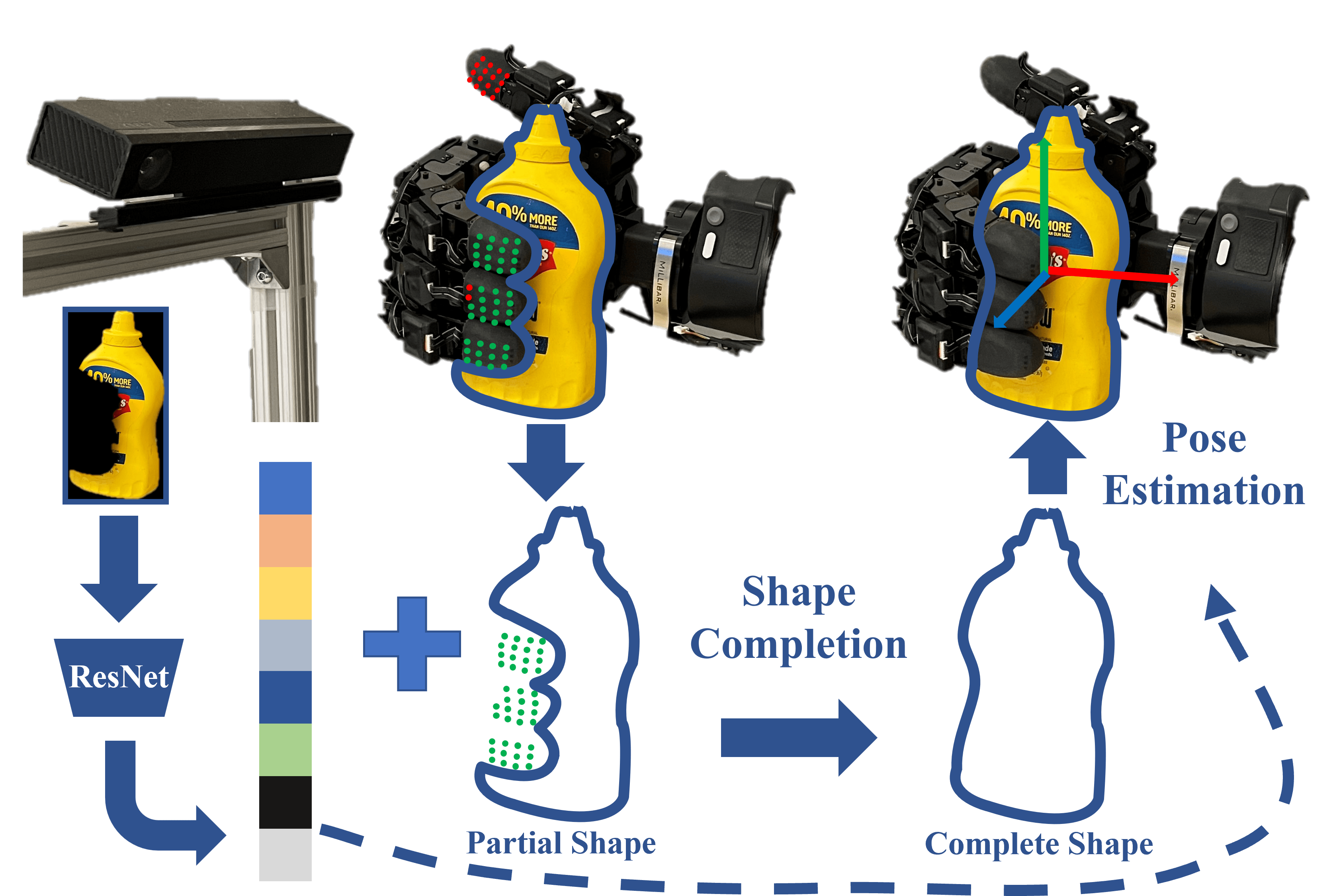}
    \caption{A high-level overview of the proposed framework. The green dots represent taxels of the tactile sensors that are in contact. RGB image and depth map are retrieved from an RGB-D sensor for object segmentation and visual feature extraction using ResNet~\cite{he_deep_2016}. The visual features, tactile data, and partial shape observation (from the vision sensor) are fed into our shape completion module and transformed into a complete shape of the in-hand object. Finally, the completed shape (in latent space) is used to estimate the object's 6D pose through a joint optimization of both shape and pose.}
    \label{fig: teaser}
\end{figure}

\IEEEPARstart{A}{n} accurate 6D pose is a de facto fundamental assumption in numerous applications, such as robotic manipulation~\cite{qin_one_2022, chen_system_2022}, autonomous driving \cite{geiger_are_2012}, and social navigation \cite{chen_socially_2017}. The absence of precise knowledge of the pose of the object makes it challenging for an agent to interact with it accurately or avoid it effectively. 
Recently, deep learning approaches have demonstrated promising results~\cite{wang_densefusion_2019, chen_epro-pnp_2022, hu_segmentation-driven_2019}. These methods, when combined with iterative refinement \cite{besl_method_1992, wang_densefusion_2019, li_deepim_2018}, leverage the object's 3D model to obtain a more accurate estimate. However, many methods do not perform well in the presence of intermediate to extreme occlusions, particularly in scenarios involving dexterous manipulation where the object is being held, grasped, or sometimes completely obscured by the robot hand. In such scenarios, finding an accurate pose of the partially observed shape is challenging.

Consequently, researchers have considered inclusion of tactile sensors into the sensor suite to improve the quality of perception during manipulation \cite{bimbo_-hand_2016, dikhale_visuotactile_2022, villalonga_tactile_2021}. \citet{villalonga_tactile_2021} leverage a template-based approach to match the visuotactile observation with the rendered shapes. \citet{dikhale_visuotactile_2022} use visuotactile data and utilize an end-to-end deep neural network and address the 6D pose estimation as a regression problem. However, they do not explicitly leverage the 3D geometry of the object. 

To this end, we introduce ViHOPE---Visuotactile In-Hand Object Pose Estimator (Fig.~\ref{fig: teaser}). ViHOPE takes visuotactile observation as input and explicitly optimizes the object shape while estimating the pose. We hypothesize that jointly optimizing the shape and pose of the object will provide more accurate estimates of the 6D pose of the object. Additionally, during deployment, by providing an estimate of the complete shape of the in-hand object, it increases explainability and potentially expands the range of applications, such as grasping \cite{varley_shape_2017}. Specifically, we first train an autoencoder to capture the geometry prior of the object and encode the object shape into a latent space. We then leverage a GAN to transfer the latent code from the partial shape space to that of the complete shape. We then use the estimated complete latent code, and the visual feature to estimate the 6D pose.

We conduct experiments on a synthetic dataset by \citet{dikhale_visuotactile_2022} and a physical robot platform. We evaluate the performance of ViHOPE on two tasks: shape completion and pose estimation. In the shape completion task, we show improved performance by a large margin compared with the prior work \cite{watkins-valls_multi-modal_2019}, where our model faithfully reconstructs the complete shape even under heavy occlusion. In the pose estimation task, we demonstrate our model outperforms the state-of-the-art visuotactile pose estimator \cite{dikhale_visuotactile_2022}. We also present results of ablation studies, in which we remove the shape completion module to confirm the effectiveness and robustness of our approach, which explicitly optimizes shape.

In summary:
1) We propose a novel visuotactile shape completion module based on the volumetric representation that accurately recovers the complete shape of the object under heavy occlusion.
2) We present a novel framework for visuotactile 6D object pose estimation that optimizes the shape completion and the pose estimation modules jointly, leveraging object geometry to improve the pose estimation.

In this letter, for simplicity, unless otherwise indicated, we use the phrase \emph{pose estimation} for visuotactile instance-level in-hand 6D object pose estimation and the phrase \emph{shape completion} for visuotactile shape completion.

\section{Relaetd Works}
In this section, we briefly summarize the related literature from two aspects: shape completion and 6D pose estimation.

\subsection{Shape Completion}

The objective of the shape completion task is to estimate the complete shape of an object from a partial observation. 
\citet{wu_multimodal_2020} approach the shape completion problem using cGAN. \citet{zhang_unsupervised_2021} formulate the problem using GAN inversion. However, their method has a time complexity that is $3,500$ times greater than direct methods \cite{cai_learning_2022}, rendering it impractical for real-time robotic applications. 

In robotic manipulation research, to counter self-occlusion and occlusions in a cluttered environment, prior works utilize shape completion in a modular \cite{varley_shape_2017}, and in an end-to-end manner \cite{jiang_synergies_2021}. \citet{wang_3d_2018} reconstruct object shape using RGB image followed by shape refinement using tactile data. \citet{watkins-valls_multi-modal_2019} complete the object's shape using visuotactile data using CNNs; however, their approach is not validated on in-hand objects.

\subsection{6D Pose Estimation}
In the field of 6D pose estimation, the problem is approached by combining various modalities, including but not limited to vision, and tactile. In this section, we will present the literature on this subject by categorizing it into two broad categories: non-visuotactile-based and visuotactile-based. 

\textbf{Non-visuotactile-based:}
The task of 6D object pose estimation has been centered around geometry. One popular trend is to find the correspondence between the observation and the 3D model of the object, such as 2D-3D correspondence \cite{park_pix2pose_2019, wang_normalized_2019, wang_gdr-net_2021, chen_epro-pnp_2022} using PnP/RANSAC \cite{fischler_random_1981} or 3D-3D correspondence \cite{li_deepim_2018} using registration methods like ICP \cite{besl_method_1992}. In these works, the exact 3D model of the object is provided during both the training and inference time. \citet{park_pix2pose_2019} and \citet{wang_normalized_2019} detect the 2D pixel-wise normalized coordinate map of the object and match it with the object's 3D model using PnP/RANSAC. While these approaches exhibit impressive performance in estimating pose on novel instances, the non-differentiable correspondence matching prohibits end-to-end training with downstream tasks. To counter this limitation, later works \cite{chen_epro-pnp_2022, wang_gdr-net_2021} investigate differentiable correspondence matching and learn the pose in an end-to-end manner. 

On the other hand, numerous works use the end-to-end method to estimate the 6D pose without explicitly finding correspondence. One of the pioneering works, PoseCNN \cite{xiang_posecnn_2018} estimates the 6D pose from monocular RGB input. \citet{gao_6d_2020} estimates 6D pose using only point cloud, leveraging PointNet \cite{qi_pointnet_2017} to extract geometry features from point cloud. DenseFusion \cite{wang_densefusion_2019} utilizes both RGB and point cloud data and fuse the extracted features \cite{he_deep_2016, qi_pointnet_2017} on a pixel-to-pixel basis to estimate the pose. The above-mentioned approaches demonstrate promising improvement on established benchmarks like LINEMOD \cite{hinterstoisser_multimodal_2011}, or YCB-Video \cite{xiang_posecnn_2018}. However, their performance has not been evaluated under heavy occlusion typically found in in-hand manipulation.

\textbf{Visuotactile-based:}
Several studies utilize both visual and tactile modalities to address the challenge of occlusions in in-hand objects.
\citet{villalonga_tactile_2021} employ a template-based approach to estimate the pose of in-hand objects by rendering a set of shape images from a 3D model in various random poses, then the recorded shape by the tactile sensor is compared with the collection to determine the most probable pose. %
\citet{dikhale_visuotactile_2022} regress the RGB-D and tactile data to a 6D pose in an end-to-end manner and outperform the prior RGB \cite{xiang_posecnn_2018} and RGB-D \cite{wang_densefusion_2019} approaches. However, they do not explicitly leverage the geometry of the object.

Unlike prior works,  ViHOPE takes advantage of both the end-to-end approach and the object geometry. We focus on jointly optimizing the object shape and the pose estimation to improve the accuracy of the pose estimate.

\begin{figure*}[ht!]
    \vspace*{0.15cm}
    \centering
        \includegraphics[width=.9\textwidth]{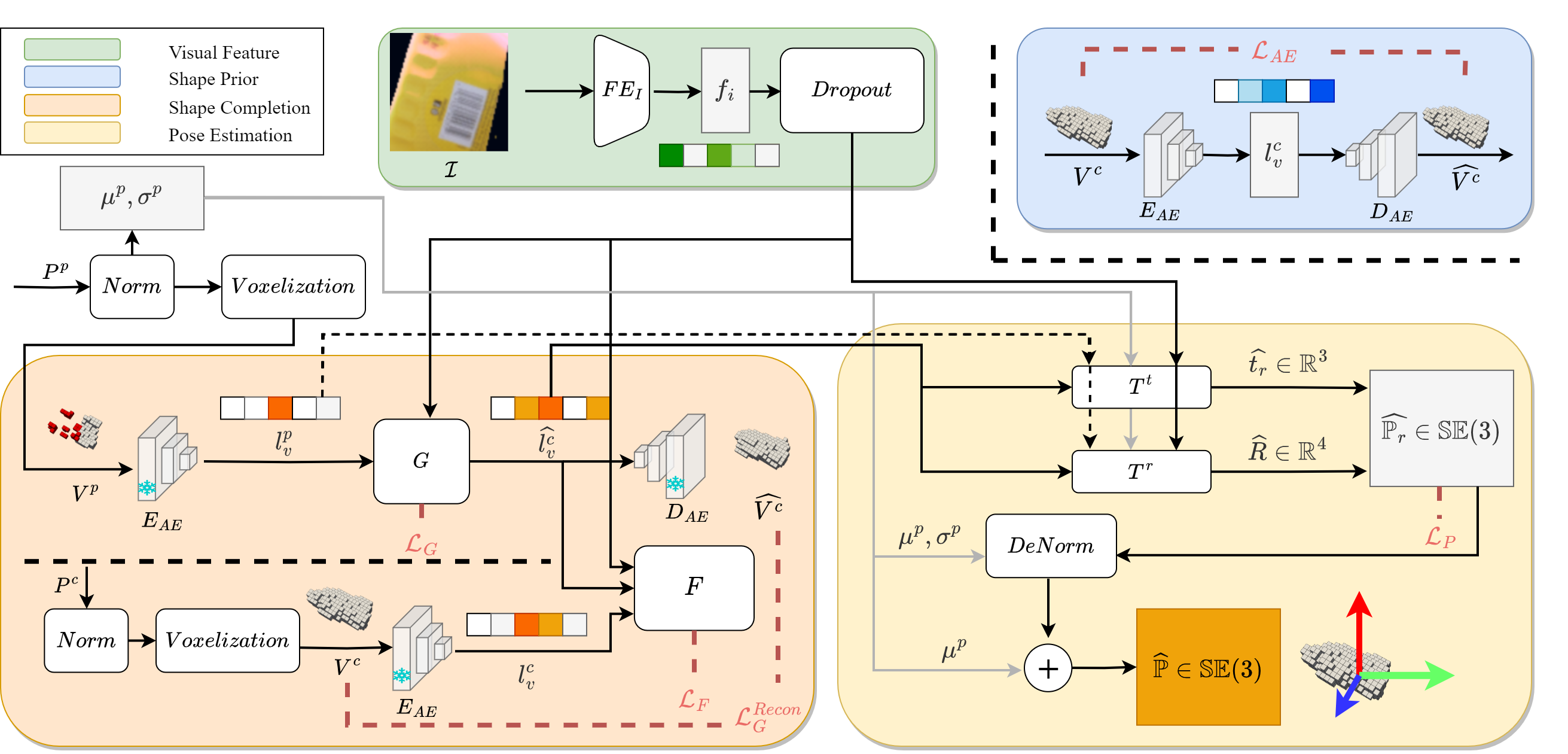}
    \caption{
    The proposed framework consists of three phases. Phase one trains the autoencoder (the blue box) in isolation to learn a shape prior. Phase two trains the shape completion module (the orange box) with frozen weights for $E_{AE}$ and $D_{AE}$ from phase one. Phase three uses the completed shape (in latent space) to estimate the object's pose in an end-to-end manner.
    }
    \label{fig: pose-network}
\end{figure*}

\section{Methodology}

Given an image $\mathcal{I}$ and its respective depth map $\mathcal{D}$, an object $\mathcal{O}$, and the tactile feedback $\mathcal{T}$ from the robot hand, our goal is to estimate the 6D pose $[R|t]$ of the object $\mathcal{O}$ with respect to the camera, where $R$ represents the 3D rotation, and $t$ represents the 3D translation. We assume the 3D model of the object $\mathcal{O}$ is available during training.

Unlike vision sensors, tactile sensors are less developed~\cite{yamaguchi_recent_2019}. There is no standard representation format for tactile data. To ensure the robustness of pose estimation algorithms against variations in tactile sensor types, we adopt the approach proposed by \citet{dikhale_visuotactile_2022}, in which the tactile feedback $\mathcal{T}$ is presented to the algorithm in the form of object-surface point cloud $P^\mathcal{T}$. That is, when the tactile sensors make contact with the object, the robot's kinematic is used to get the 3D position of each taxel in contact with the object, using a user-defined threshold on the sensor feedback. 

The proposed visuotactile pose estimation framework (Fig.~\ref{fig: pose-network}) consists of: a visuotactile shape completion module and a pose estimation module. The shape completion module explicitly optimizes the object's shape from a partial observation of the object from the vision and tactile sensors (Section \ref{sec: shape-completion-method}). The pose estimation module uses the output of the shape completion module, in the form of latent code, along with the visual features and point cloud normalization scalars to estimate the 6D pose of the object (Section \ref{sec: pose-estimation-method}).

\subsection{Shape Completion}
\label{sec: shape-completion-method}
Our framework starts from a volumetric shape completion module, which consists of two steps: 1) learning a shape prior from the object model, i.e., full observation, 2) recovering the complete shape by learning a mapping from partial observation to a complete one, in the latent space.

\subsubsection{Learning the shape prior}
We first train an autoencoder (the blue box in Fig.~\ref{fig: pose-network}) to capture the shape prior of the object. To capture the prior under different orientations, we apply random $SO(3)$ rotations to the object $\mathcal{O}$, and voxelize it to form an augmented dataset. The autoencoder consists of two components: an encoder $E_{AE}$ and a decoder $D_{AE}$. The encoder encodes the input voxel occupancy grid $V \in \mathbb{R}^{n_x \times n_y \times n_z}$ into a latent code $l_v \in \mathbb{R}^{n_l}$ using a set of 3D convolutional layers. The decoder recovers the original voxel occupancy grid from the latent code  as $\widehat{V} \in \mathbb{R}^{n_x \times n_y \times n_z}$ using symmetrical 3D deconvolutional layers. We apply a batch normalization layer after each layer, followed by a ReLU activation function. We set $n_x = n_y = n_z = 32$ and $n_l = 128$, empirically, and optimize the autoencoder model using the Jaccard index loss
\begin{equation}
    \mathcal{L}_{AE} = 1 - \frac{|V^c \cap D_{AE}(E_{AE}(V^c))|}{|V^c \cup D_{AE}(E_{AE}(V^c))|} .
    \label{eqn:jaccard_loss}
\end{equation}

Note, higher voxel grid resolution can improve accuracy but comes with increased computation and memory costs. Some works have explored methods to mitigate these costs~\cite{tatarchenko_octree_2017, li2023stereovoxelnet}; however, such optimization techniques are beyond the scope of this work.

\subsubsection{Recovering the complete shape}
After training the autoencoder to convergence, we optimize the entire shape completion model (the orange box in Fig.~\ref{fig: pose-network}). The encoder $E_{AE}$ and the decoder $D_{AE}$ of the shape completion module are initialized with the weights from the previous step and are frozen in this step~\cite{wu_multimodal_2020}. 

The inputs to the shape completion model consist of occluded observational data, that is, a partial visuotactile point cloud $P^p \in \mathbb{R}^{3 \times N}$ and an RGB image $\mathcal{I} \in \mathbb{R}^{H \times W \times 3}$. We first obtain a semantic segmentation mask $\mathcal{S}$ of the object $\mathcal{O}$ using the RGB image $\mathcal{I}$. %
We then segment the depth map $\mathcal{D}$ using the mask $\mathcal{S}$ as $\mathcal{D}^p$ and transform $\mathcal{D}^p$ to its respective point cloud form $P^\mathcal{D}$. Combining the point cloud $P^\mathcal{D}$ with the tactile point cloud $P^\mathcal{T}$, we obtain the observation $P^p = P^\mathcal{D} \cup P^\mathcal{T}$ of the object $\mathcal{O}$. 
To improve the training efficiency, we first normalize the input partial point cloud $P^p$ using its centroid $\mu^p \in \mathbb{R}^{3}$ and the farthest distance from the centroid $\sigma^p \in \mathbb{R}$. The normalized point cloud is voxelized as $V^p$ and encoded using the frozen encoder $E_{AE}$ into a partial latent vector $l_v^p$ in the partial latent space $\mathcal{M}^p$.
During training, we use the ground-truth complete voxel grid $V^c$ to calculate the losses.

To recover the complete shape from the partial observation, we seek a mapping for shape latent code from the partial latent space to the complete latent space $\mathcal{M}^p \mapsto \mathcal{M}^c$. Inspired by \citet{wu_multimodal_2020}, we find the mapping using a cGAN~\cite{mirza_conditional_2014}.
We condition the cGAN on the partial latent vector $l_v^p$ and the visual feature $f_i$, which is extracted from the input image $\mathcal{I}$ using a pretrained ResNet \cite{he_deep_2016} feature extractor $FE_I$. $FE_I$ is finetuned during the training period and regularized by a dropout layer. 
The dropout layer is only activated during training and deactivated during testing.
Therefore our generator is trained as 
$G: (\mathcal{M}^p, p(f_i)) \mapsto \mathcal{M}^c$.
We pass the estimated complete latent vector $\widehat{l^c_v} \in \mathcal{M}^c$ from the generator to the discriminator $F$ along with the ground-truth complete latent vector $l_v^c$, obtained by applying the same procedure on the ground-truth complete point cloud $P^c$. Having the same conditioning as $G$, the discriminator $F$ is trained as a binary classifier to distinguish the real complete latent vector $l_v^c$ and the fake complete latent vector $\widehat{l^c_v}$. At the end, we feed the estimated complete latent vector $\widehat{l^c_v}$ into the frozen decoder $D_{AE}$ to reconstruct the complete shape $\widehat{V^c}$. 

The shape completion model is optimized using three losses: the discriminator loss $\mathcal{L}_F$, the generator loss $\mathcal{L}_G$, and the reconstruction loss $\mathcal{L}_G^{Recon}$. The discriminator loss penalizes the discriminator if it can't distinguish the real and fake latent vectors. On the other side of this min-max game, the generator loss encourages the generator to fool the discriminator. We leverage the loss functions from LSGAN \cite{mao_least_2017} to stabilize the training

\begin{equation}
\begin{split}
    \mathcal{L}_{F} & = \mathbb{E}_{V^c, \mathcal{I}} [F_{cGAN}(E_{AE}(V^c), FE_I(\mathcal{I})) - 1]^2 \\
    & + \mathbb{E}_{V^p, \mathcal{I}}[F_{cGAN}(G_{cGAN}(E_{AE}(V^p), FE_I(\mathcal{I})))]^2
    \label{eqn:discriminator_loss}
\end{split}
\end{equation}

\begin{equation}
\begin{split}
   \mathcal{L}_{G} & = \mathbb{E}_{V^p, \mathcal{I}}[F_{cGAN}(G_{cGAN}(E_{AE}(V^p), FE_I(\mathcal{I}))) - 1]^2 .
    \label{eqn:generator_loss}
\end{split}
\end{equation}

To further stabilize the GAN training and guide the model, we include the reconstruction loss $\mathcal{L}_G^{Recon}$ that directly measures the differences between the ground-truth complete shape $V^c$ and the estimated complete shape $\widehat{V^c} \triangleq D_{AE}(G_{cGAN}(E_{AE}(V^p), FE_I(\mathcal{I})))$ using Jaccard index loss

\begin{equation}
   \mathcal{L}_G^{Recon} = 1 -
   \frac{|V^c \cap \widehat{V^c}|}{|V^c \cup \widehat{V^c}|}.
    \label{eqn:recon_loss}
\end{equation}

Therefore, the overall training objective for our shape completion model is
\begin{equation}
   \argmin_{G, FE_I} \argmax_{F} \mathcal{L}_{F} + \mathcal{L}_{G} + \alpha \mathcal{L}_G^{Recon},
    \label{eqn:total_shape_comp_loss}
\end{equation}
where $\alpha$ is the weight for the reconstruction loss.

\subsection{Pose Estimator}
\label{sec: pose-estimation-method}

We use two simple four-layer MLP models to estimate the pose of the object.
Our pose estimators $T^t$ and $T^r$ take the estimated complete shape latent code $\widehat{l^c_v}$, visual features $f_i$, normalization factors $\mu^p$ and $\sigma^p$, and a skip connection from partial latent $l_v^p$ as input and estimate the 3D translation residual $\widehat{t_r} \in \mathbb{R}^3$ and 3D rotation in quaternion form $\widehat{R} \in \mathbb{R}^4$, respectively. Note, similar to the prior works~\cite{wang_densefusion_2019, dikhale_visuotactile_2022}, instead of estimating the absolute translation $t$, we estimate the residual of the translation $t_r=t-\mu^p$. We use the residual pose $\widehat{\mathbb{P}_r} = [\widehat{R} | \widehat{t_r}]$ to calculate the point cloud loss $\mathcal{L}_P$ \cite{wang_densefusion_2019}:

\begin{equation}
    \mathcal{L}_P = \frac{1}{k} \sum_{x \in \mathcal{K}} ||(Rx+t_r) - (\widehat{R}x+\widehat{t_r})  ||,
\end{equation}

where $\mathcal{K}$ denotes a set of points randomly sampled from the object's 3D model, and $k$ represents the cardinality $|\mathcal{K}|$. The point cloud loss minimizes the distance between the points on the ground-truth pose and their respective points on the models transformed using the estimated pose. The overall loss function is shown in Equation \ref{eqn:total_loss},
\begin{equation}
   \argmin_{G, FE_I, T^t, T^r} \argmax_{F} \mathcal{L}_{F} + \mathcal{L}_{G} + \alpha \mathcal{L}_G^{Recon} + \beta \mathcal{L}_P ,
    \label{eqn:total_loss}
\end{equation}
where $\beta$ is the weight for the point cloud loss. 

To speed up convergence, our method is trained in four steps: i) The autoencoder ($E_{AE}+D_{AE}$) is trained in isolation, and weights are frozen; ii) The shape completion module ($G+F$) is trained; iii) The pose estimator ($T^t+T^r$) is trained while freezing the shape completion module; iv) the shape completion module in unfrozen and trained end-to-end ($G+F+T^t+T^r$), similar to \cite{liang_sscnav_2021, gervet_navigating_2022}. 
\begin{figure*}[ht!]
    \vspace*{0.15cm}
    \centering

    \begin{subfigure}{.24\textwidth}
        \centering
        \includegraphics[width=3.4cm, height=3.4cm]{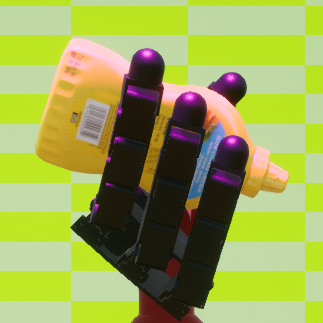}
    \end{subfigure}
    \begin{subfigure}{.24\textwidth}
        \centering
        \includegraphics[width=3.4cm, height=3.4cm]{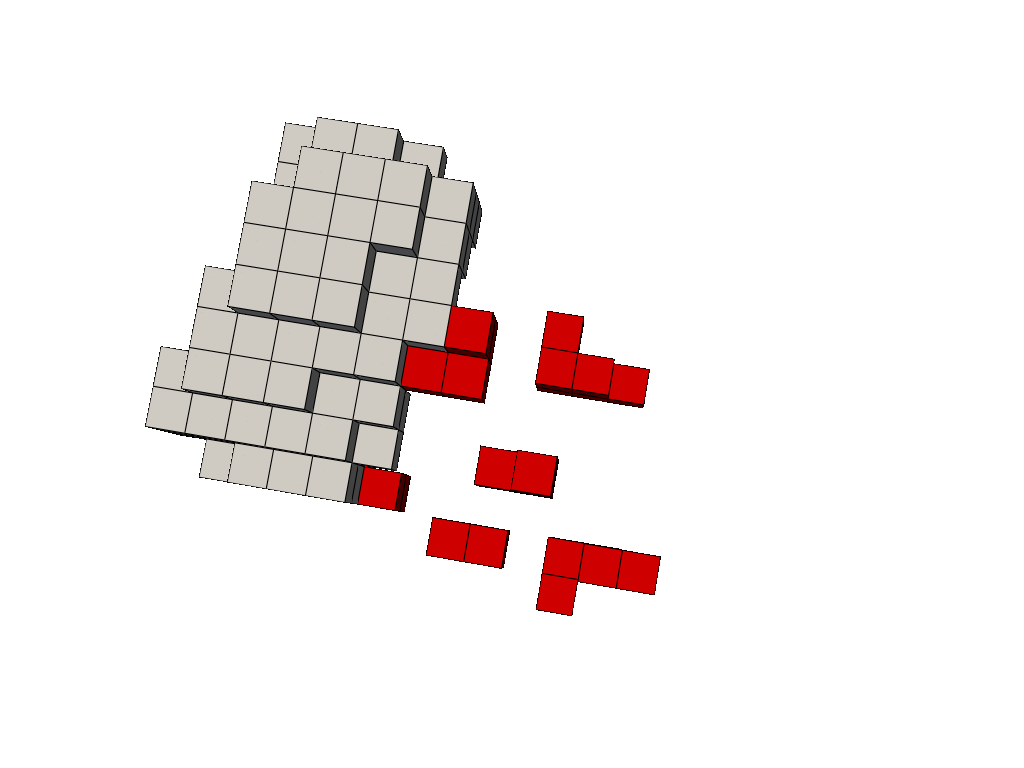}
    \end{subfigure}
    \begin{subfigure}{.24\textwidth}
        \centering
        \includegraphics[width=3.4cm, height=3.4cm]{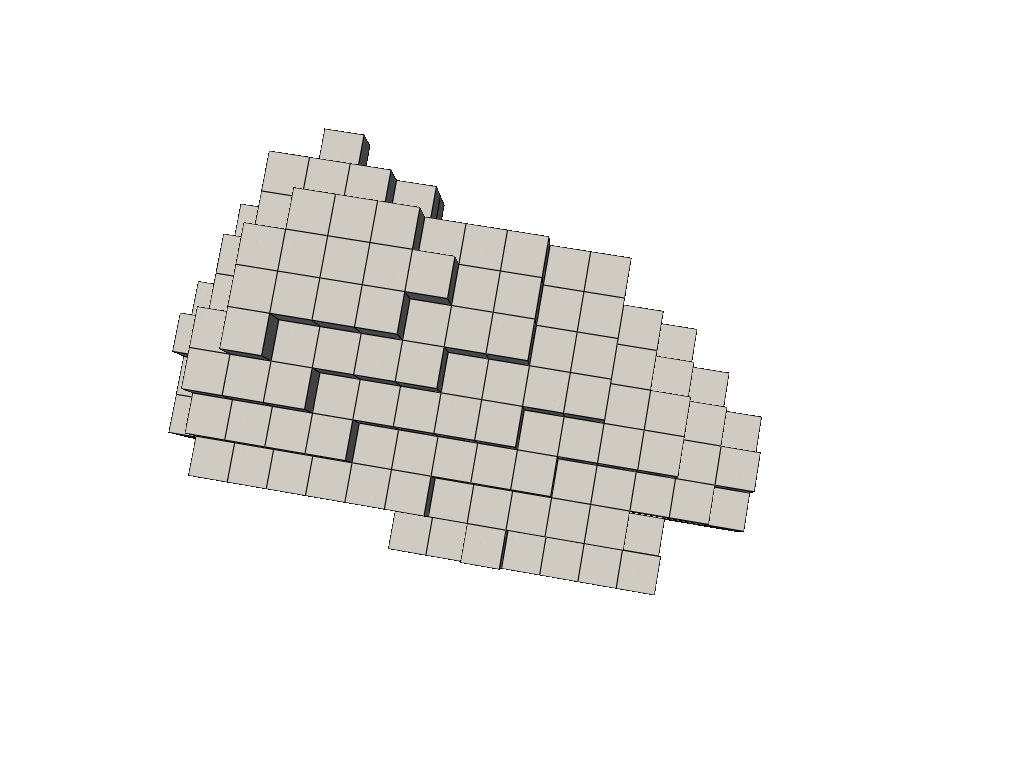}
    \end{subfigure}
    \begin{subfigure}{.24\textwidth}
        \centering
        \includegraphics[width=3.4cm, height=3.4cm]{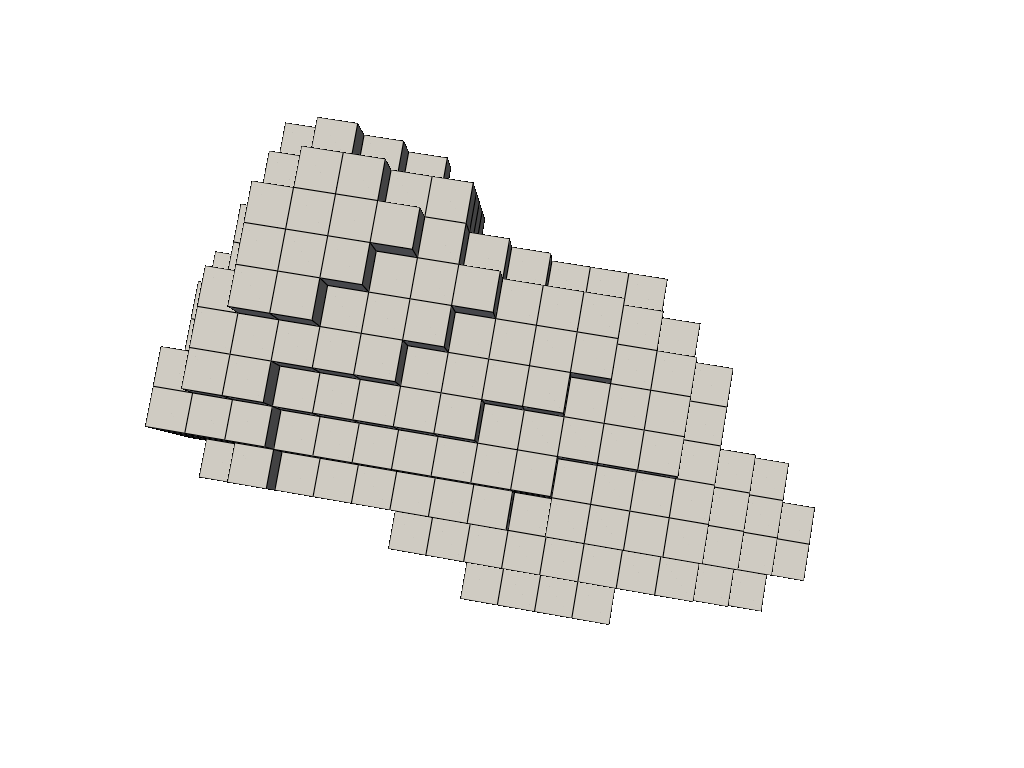}
    \end{subfigure}

    \begin{subfigure}{.24\textwidth}
        \centering
        \includegraphics[width=3.4cm, height=3.4cm]{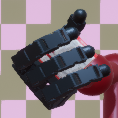}
    \end{subfigure}
    \begin{subfigure}{.24\textwidth}
        \centering
        \includegraphics[width=3.4cm, height=3.4cm]{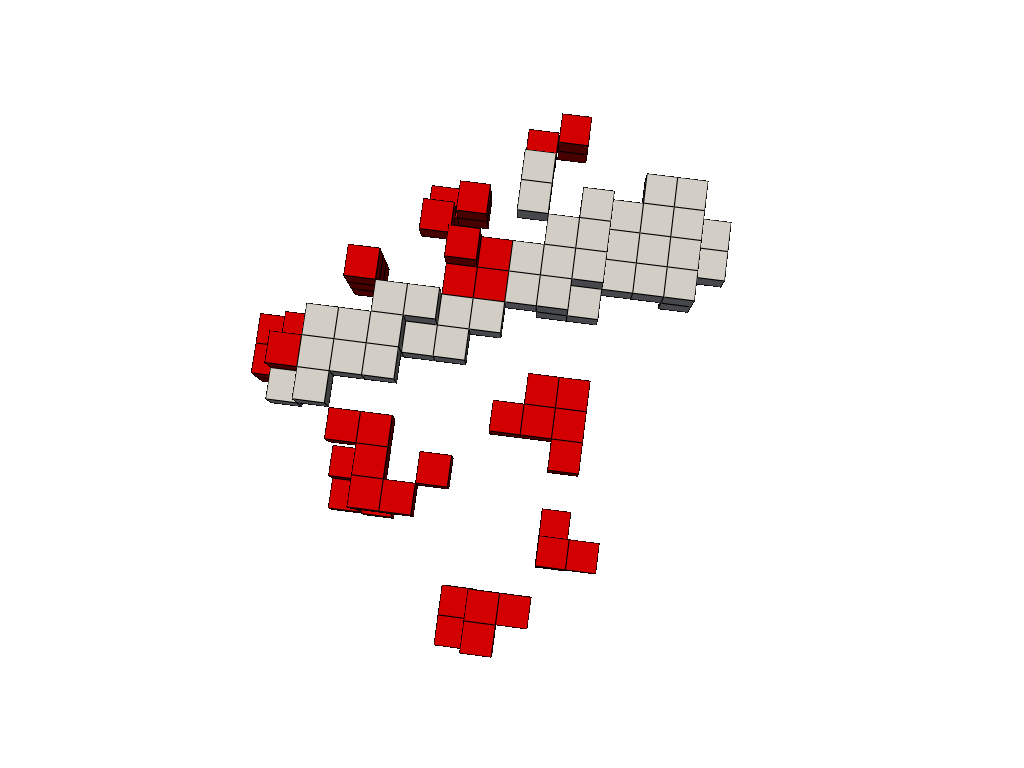}
    \end{subfigure}
    \begin{subfigure}{.24\textwidth}
        \centering
        \includegraphics[width=3.4cm, height=3.4cm]{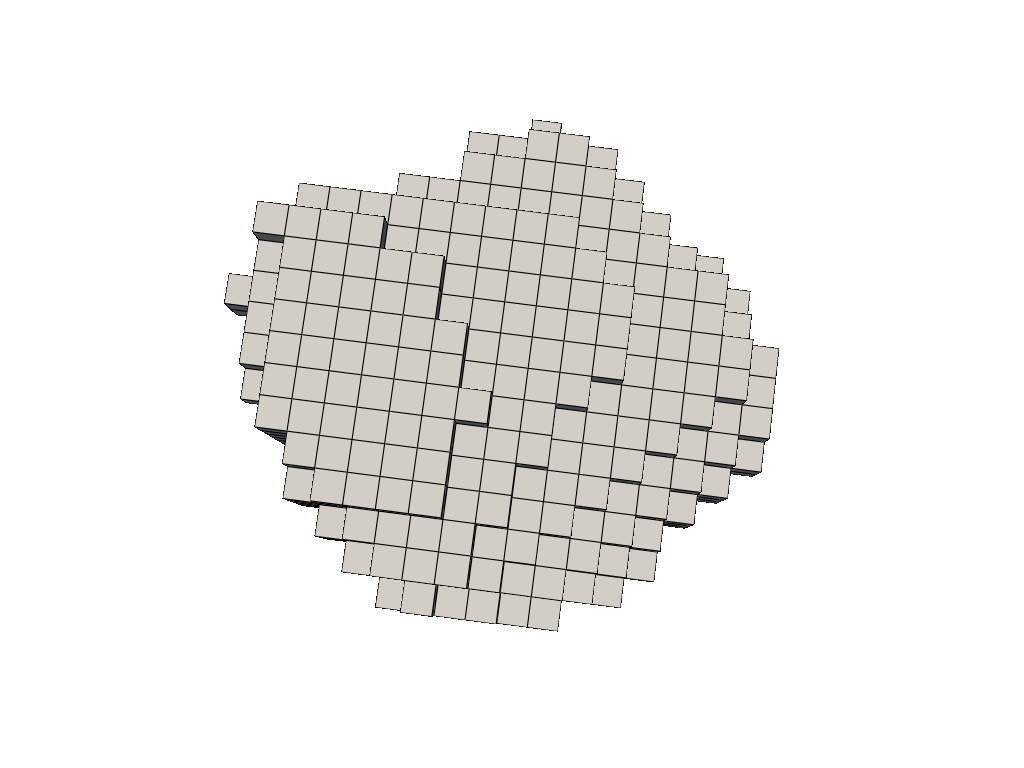}
    \end{subfigure}
    \begin{subfigure}{.24\textwidth}
        \centering
        \includegraphics[width=3.4cm, height=3.4cm]{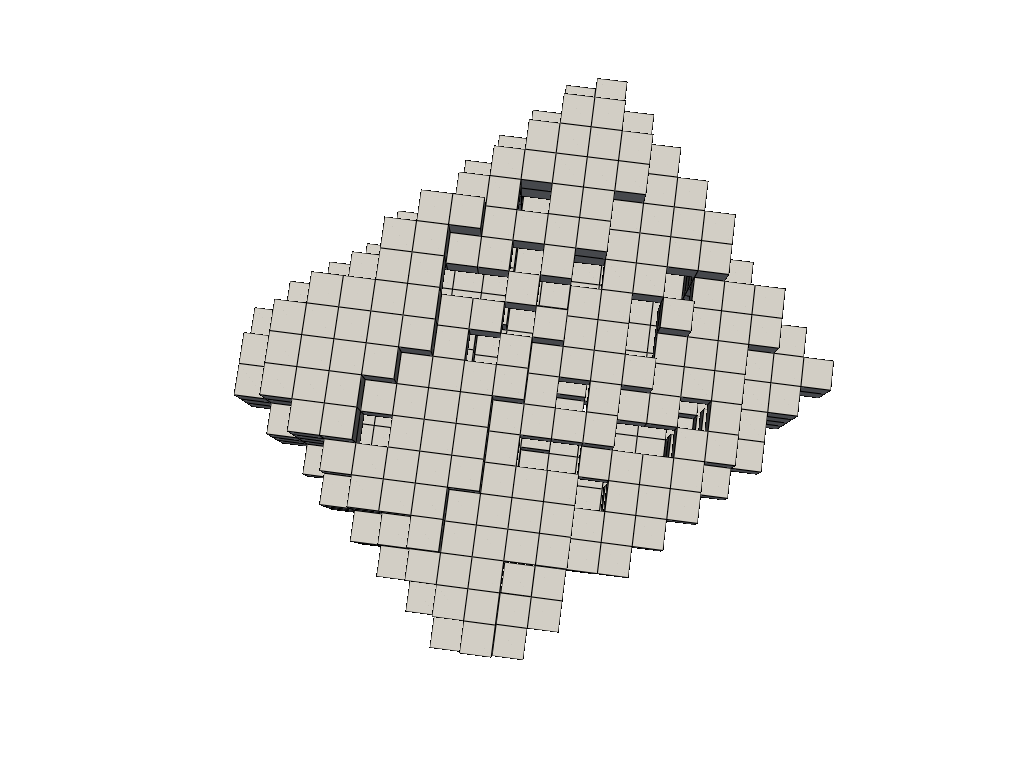}
    \end{subfigure}

    \subcaptionbox{Camera View}%
  [.24\textwidth]{
    \includegraphics[width=3.4cm, height=3.4cm]{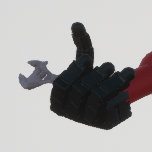}
  }
    \subcaptionbox{Partial Shape}%
  [.24\textwidth]{
    \includegraphics[width=3.4cm, height=3.4cm]{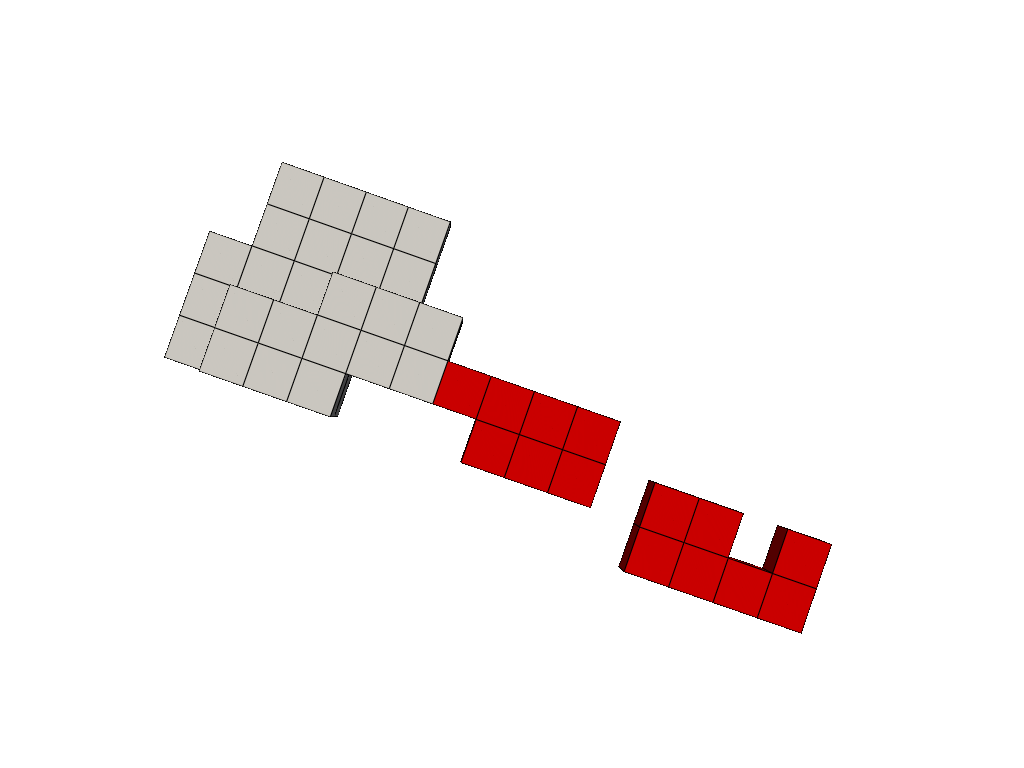}
  }
    \subcaptionbox{Completed Shape}%
  [.24\textwidth]{
    \includegraphics[width=3.4cm, height=3.4cm]{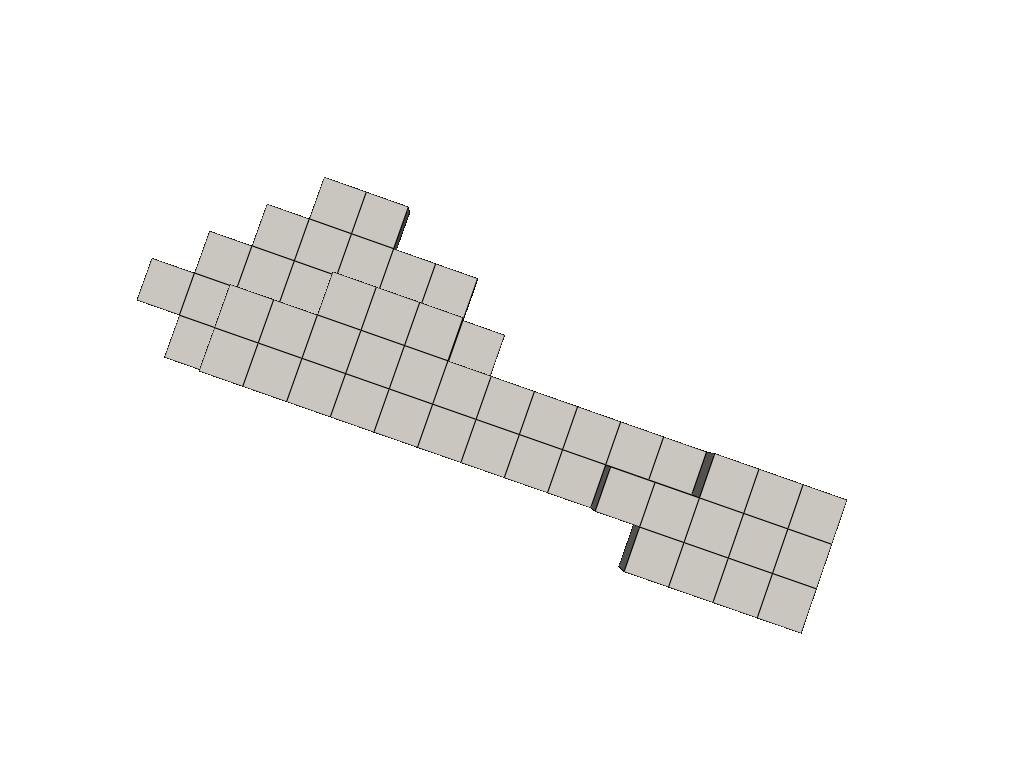}
  }
   \subcaptionbox{Ground Truth}%
  [.24\textwidth]{
    \includegraphics[width=3.4cm, height=3.4cm]{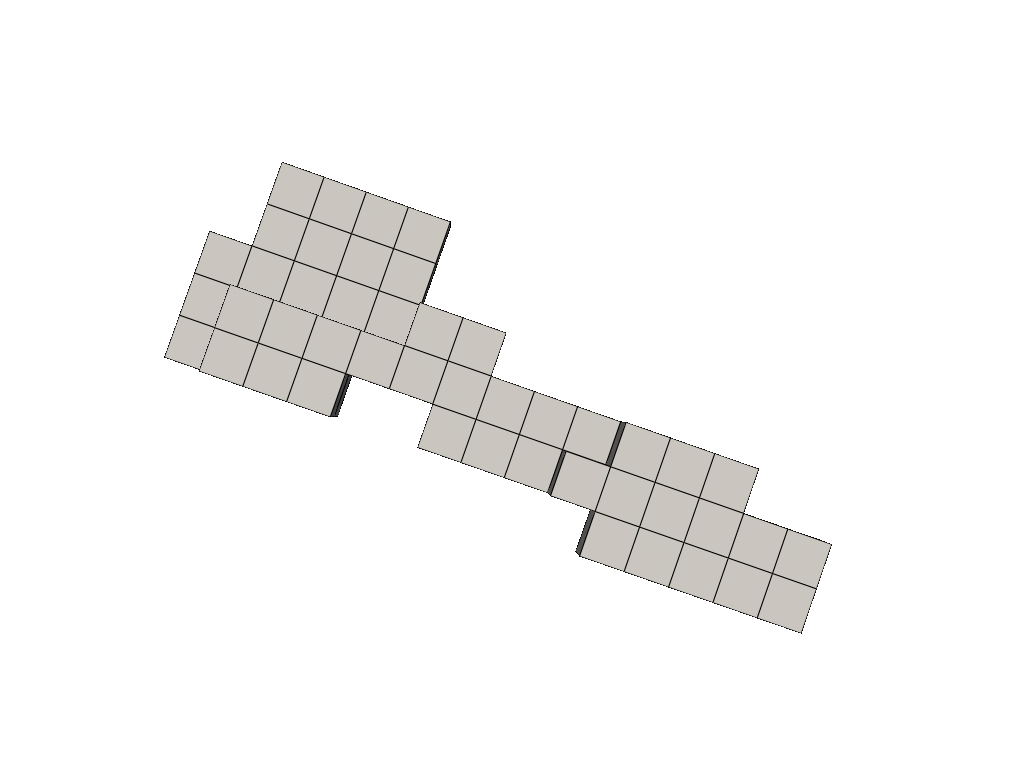}
  }

    \caption{Visuotactile shape completion results. The gray and red voxel represent RGB-D and tactile observations, $P^\mathcal{D}$ and $P^\mathcal{T}$, respectively. From left to right, we show the cropped RGB image from the main camera, the observed partial shape of the in-hand object, the completed shape by our approach, and the ground-truth shape. From top to bottom are the samples drawn from the YCB categories ``006\_mustard\_bottle", ``009\_gelatin\_box", and ``044\_adjustable\_wrench".}
    \label{fig: shape-completion}
\end{figure*}

\section{Experiments}

Our experiments are designed to assess the efficacy of our proposed framework: 1) in the level of accuracy achieved by our shape completion module in reconstructing object shapes; 2) the impact of explicit shape completion on the quality of 6D pose estimation; 3) the contribution of each component of the framework to pose estimation accuracy; and 4) the sensitivity of the framework to variations in occlusion levels and tactile contact points. This section outlines our experimental setup including model training and performance evaluation. The model was trained and tested on a synthetic dataset, and then transferred to a real physical robot to study the framework's robustness in sim-to-real transfers.
 
\paragraph*{Synthetic Dataset}

We use VisuoTactile synthetic dataset from \citet{dikhale_visuotactile_2022} to train our framework. In this dataset, a subset of 11 YCB objects \cite{calli_ycb_2015} are selected based on their graspability. A total number of $20$K distinct in-hand poses are simulated per object. In particular, Unreal Engine 4.0 has been used to render photo-realistic observational data of a 6~DoF robot arm with a 4-fingered gripper equipped with 12 32x32 tactile sensors~(3~per~finger). A main RGB-D camera captures images of the robot holding an object. Each tactile sensor captures object surface contact points in a point cloud format. Each data sample is generated by randomizing the in-hand object pose, the robot fingers configuration, and the robot arm orientation and position. Domain randomization is also applied for the color and pattern of the background and workspace desk.

\paragraph*{Real Robot}
We test our model on the Allegro Hand (Wonik Robotics) and the Sawyer robot arm (Rethink Robotics). Our Allegro Hand is instrumented with the uSkin 4x4 tactile sensors and uSkin Curved tactile sensors from XELA Robotics. Each finger has three 4x4 tactile sensors (16 taxels) and one Curved tactile sensor (30 taxels). We use a Microsoft Kinect V2 camera to collect RGB-D data.

\paragraph*{Implementation Details}
We utilize an ImageNet pre-trained ResNet-34 model as our visual feature extractor $FE_I$. We apply center crop, Gaussian blur, and color jitter augmentations to the input image. We implement our pose estimators $T^r$ and $T^t$ using four-layer MLPs with layers of $[512, 256, 32, 4]$ and $[512, 256, 32, 3]$, respectively. We trained our autoencoder for 500 epochs, shape completion module for 1000 epochs, pose estimator for 1000 epochs, and the entire end-to-end model for 2000 epochs using the Adam optimizer with a learning rate of $1\times10^{-3}$, $5\times10^{-4}$, $1\times10^{-3}$, and $5\times10^{-5}$, respectively. We set the reconstruction loss $\alpha$ and the point cloud loss $\beta$  (Equation \ref{eqn:total_loss}) to $30$ and $5000$, respectively.

\section{Results}
\label{sec:results}

\subsection{Shape Completion}

\subsubsection{Qualitative Result}
Fig.~\ref{fig: shape-completion} represents a visualization of the input and output of our shape completion module, which uses visuotactile observations to faithfully complete the shape of the in-hand object. As the 3D reconstruction can have varying levels of occlusion from different viewing angles, we recommend viewing the accompanying video for a clearer and improved visualization.

\subsubsection{Quantitative Result}

\begin{table}[t!]
 \caption{Quantitative shape completion result.}
 \label{tab: shape-completion}
\centering
 \begin{tabular}{l | c | c } 
 \hline
 Method & IoU $\uparrow$ & CD $\downarrow$  \\ 
 \hline
 \citet{watkins-valls_multi-modal_2019} & 0.142 & 0.125 \\
  Vision Only & 0.341 & 0.042  \\
 \textbf{ViHOPE (Ours)} & \textbf{0.519} & \textbf{0.015}  \\
 \hline
 \end{tabular}
\end{table}

We compare our shape completion module with the seminal work from Watkins-Valls et al.~\cite{watkins-valls_multi-modal_2019} using two metrics: Intersection over Union (IoU) and Chamfer Distance (CD). We implement their proposed model~\cite{varley_shape_2017, watkins-valls_multi-modal_2019} using PyTorch %
with a minor modification. To ensure a fair evaluation, a consistent voxel occupancy grid resolution was used. Our implementation utilizes a $32^3$ voxel grid for input and output, instead of the  $40^3$ voxel grid utilized in the original work. This modification allowed seamless integration of the shape completion module into our established pipeline. Results in Table \ref{tab: shape-completion} show a 265.5\% increase in IoU and an 88\% decrease in CD, demonstrating the robustness of our model under challenging conditions.

We also evaluate the performance of the shape completion model by removing the tactile modality (Vision Only). The result suggests the significance of tactile modality for completing the shape of an in-hand object.

\subsection{6D Pose Estimation}

\begin{table*}[ht!]
\vspace*{0.15cm}
\caption{A comparison of our approach with the state-of-the-art is presented in the first half of the table, followed by the results of our ablation studies in the second half. The modalities used by the methods are highlighted in the second column.}
 \label{tab: modalities}
\centering
 \begin{tabular}{ l | c | c | c | r | r | r | r } 
 \hline
 \multirow{2}{*}{Method} & \multicolumn{3}{|c|}{Modalities} & \multirow{2}{*}{Position Error (cm) $\downarrow$} & \multirow{2}{*}{Angular Error (deg) $\downarrow$} & \multirow{2}{*}{ADD (cm) $\downarrow$} & \multirow{2}{*}{ADD-S (cm) $\downarrow$} \\ 
 & RGB  & point cloud & tactile &  &  & & \\ 
 \hline
 PoseCNN \cite{xiang_posecnn_2018} & \checkmark{}  & & & 6.146 $\pm$ 0.023  & 10.897 $\pm$ 0.082 & - & - \\
 DenseFusion \cite{wang_densefusion_2019} & \checkmark{}  & \checkmark{} &  & 0.640 $\pm$ 0.004 & 9.969 $\pm$ 0.117 & 1.037 $\pm$ 0.008 & 0.571 $\pm$ 0.003 \\ 
 ViTa \cite{dikhale_visuotactile_2022} & \checkmark{} & \checkmark{} & \checkmark{} & 0.299 $\pm$ 0.002 & 8.074 $\pm$ 0.105 & 0.825 $\pm$ 0.007 & 0.474 $\pm$ 0.002 \\
 \hline
 No-Shape-Completion & \checkmark{} & \checkmark{} & \checkmark{} & 0.258 $\pm$ 0.018 & 4.104 $\pm$ 0.049 & 0.400 $\pm$ 0.019 & 0.282 $\pm$ 0.018 \\
 No-Vis-GAN & \checkmark{} & \checkmark{} & \checkmark{} & 1.613 $\pm$ 0.333 & 4.132 $\pm$ 0.058 & 1.745 $\pm$ 0.333 & 1.623 $\pm$ 0.332 \\
 No-Vis-MLP & \checkmark{} & \checkmark{} & \checkmark{} &  \textbf{0.156 $\pm$ 0.001} & 5.677 $\pm$ 0.083 & 0.403 $\pm$ 0.004 & \textbf{0.214 $\pm$ 0.001} \\
 No-Tactile & \checkmark{}  & \checkmark{} &  & 1.614 $\pm$ 0.015 & 17.228 $\pm$ 0.165 & 2.023 $\pm$ 0.017 & 0.774 $\pm$ 0.009 \\
 No-Point-Cloud & \checkmark{}  &  & \checkmark{} & 0.861 $\pm$ 0.010 & 14.245 $\pm$ 0.096 & 1.478 $\pm$ 0.011 & 0.655 $\pm$ 0.009 \\
 \hline
 \textbf{ViHOPE (Ours)} & \checkmark{} & \checkmark{} & \checkmark{} &  0.194 $\pm$ 0.009 & \textbf{2.873 $\pm$ 0.036} & \textbf{0.298 $\pm$ 0.009} & \textbf{0.214 $\pm$ 0.008}\\
 \hline
 \end{tabular}
\end{table*}

\paragraph*{The Metrics}
We evaluate the performance of our pose estimator using two metrics: position error and angular error. The position error is determined as the L2 norm of the difference between the estimated translation vector and the ground truth translation vector, $||t-\hat{t}||_2$. The angular error is computed as the inverse cosine of the inner product of the estimated rotation quaternion and the ground truth quaternion, $\cos^{-1}(2 \langle R, \hat{R} \rangle^2 -1)$.

\subsubsection{Comparsion with state-of-the-art}

\begin{figure}[t!]
    \begin{subfigure}{\columnwidth}
        \centering
        \includegraphics[width=\columnwidth]{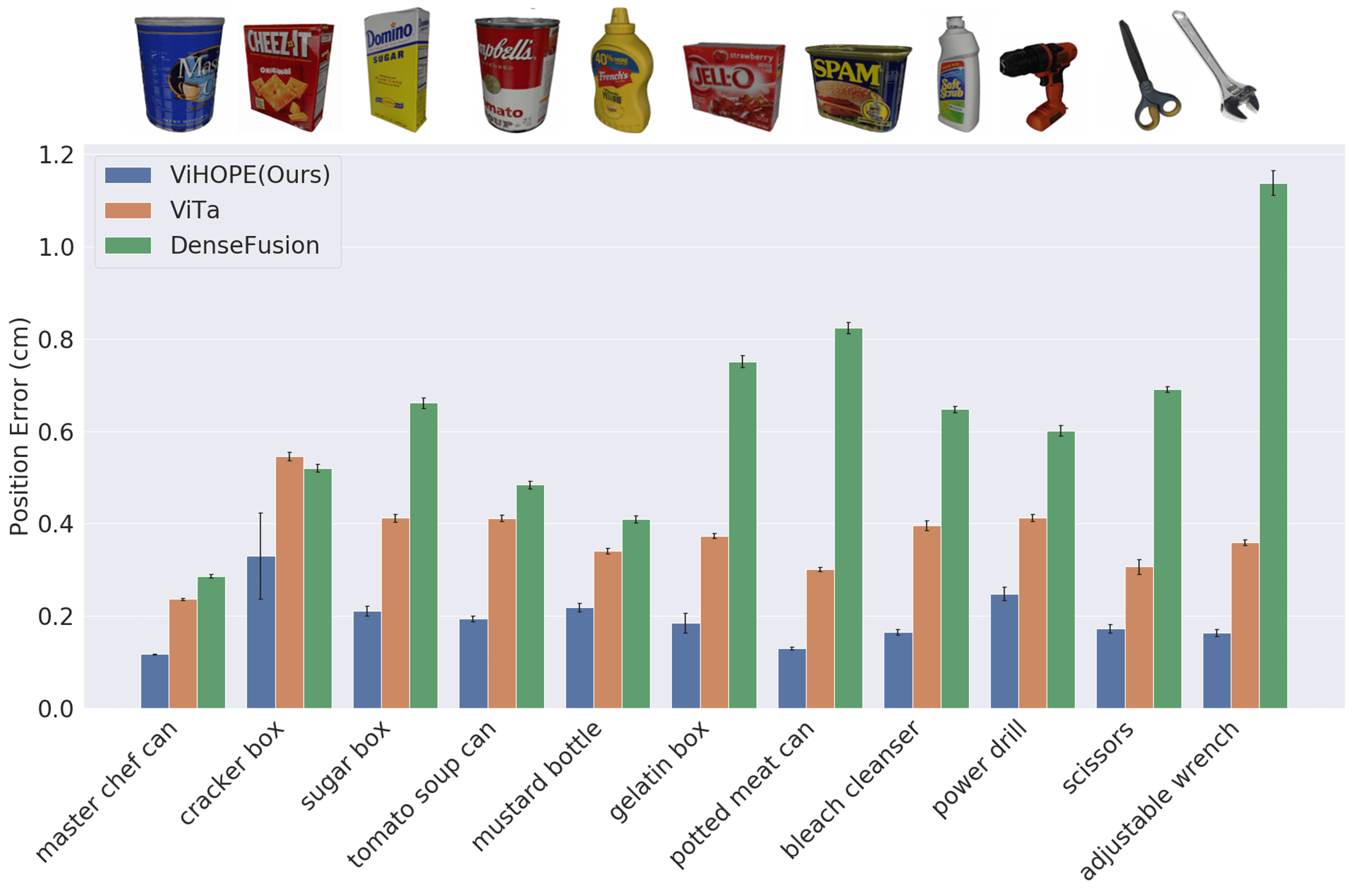}
        \hfill
    \end{subfigure}
    \begin{subfigure}{\columnwidth}
        \centering
        \includegraphics[width=\columnwidth]{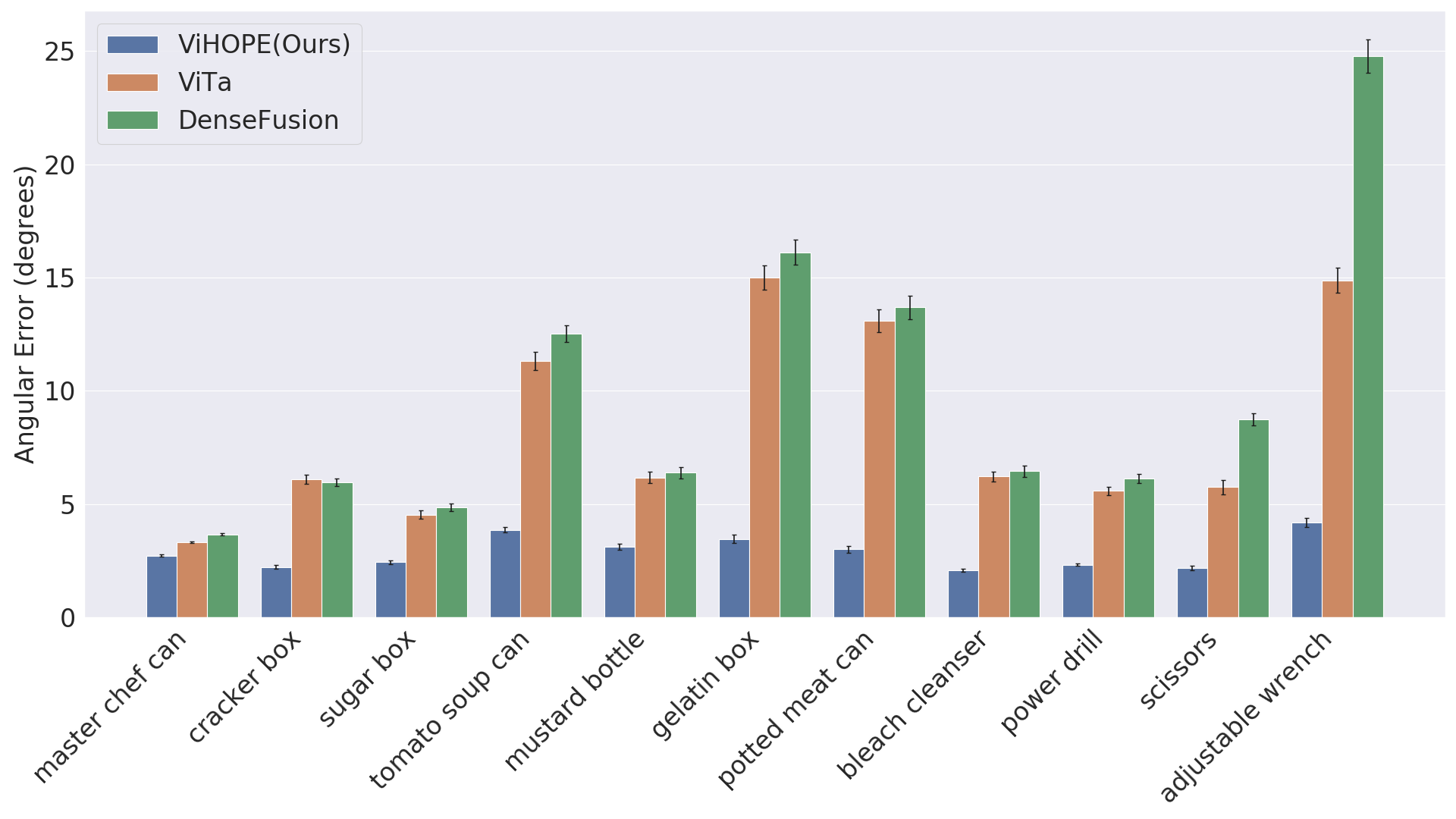}
    \end{subfigure}
    \caption{Performance comparison with the state-of-the-art.}
    \label{fig: compare-sota}
\end{figure}

We compare the performance of our pose estimation network with two seminal works: i) the visuotactile-based estimator (ViTa) \cite{dikhale_visuotactile_2022}, and ii) the RGB-D-based estimator, DenseFusion \cite{wang_densefusion_2019}. In Fig.~\ref{fig: compare-sota}, we provide a per-instance numerical analysis on 11 YCB objects.
Our approach outperforms ViTa and DenseFusion by a large margin on each object with statistical significance, suggesting explicit shape optimization is more effective compared to implicit methods.

\subsubsection{Performance under different occlusion level}
\begin{figure}[t!]
    \centering
    \begin{subfigure}{.49\columnwidth}
        \centering
        \includegraphics[width=\columnwidth]{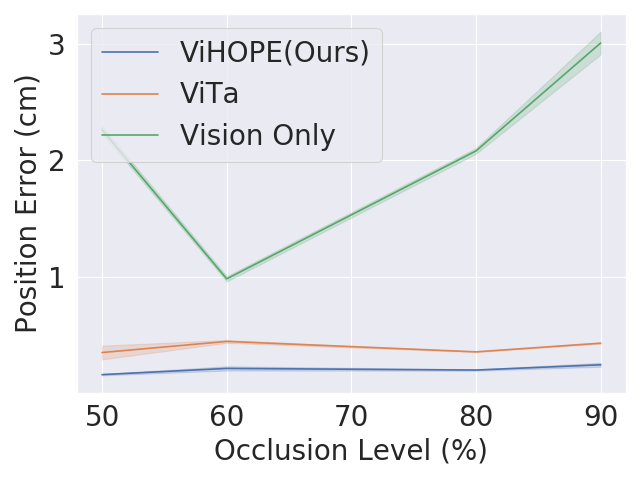}
    \end{subfigure}
    \begin{subfigure}{.49\columnwidth}
        \centering
        \includegraphics[width=\columnwidth]{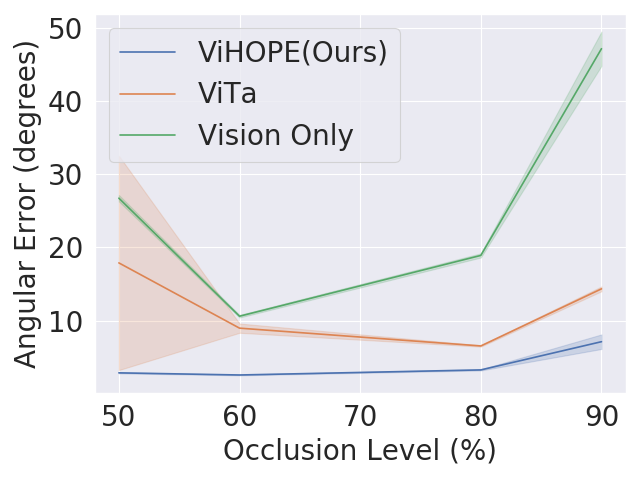}
    \end{subfigure}
    \caption{Performance under different levels of occlusion}
    \label{fig: occlusion}
\end{figure}

We evaluate the performance of our model under different levels of occlusion. The results of our evaluation are presented in Fig.~\ref{fig: occlusion}, where it can be observed that the model demonstrates a robust performance in the presence of increasing levels of occlusion. It is worth noting that our method maintains its performance as compared to ViTa, which suggests that our model is able to handle occlusion effectively and still produce competitive performance. We further evaluate the performance by removing the tactile modality (Vision Only). The result confirms the significant contribution of tactile modality under severe occlusions.

\subsubsection{Performance under different tactile points}
\begin{figure}[t!]
    \vspace*{0.15cm}
    \centering
    \begin{subfigure}{.49\columnwidth}
        \centering
        \includegraphics[width=\columnwidth]{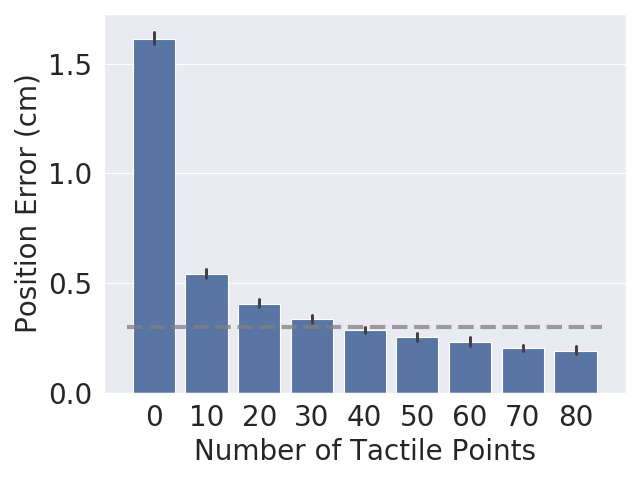}
    \end{subfigure}
    \begin{subfigure}{.49\columnwidth}
        \centering
        \includegraphics[width=\columnwidth]{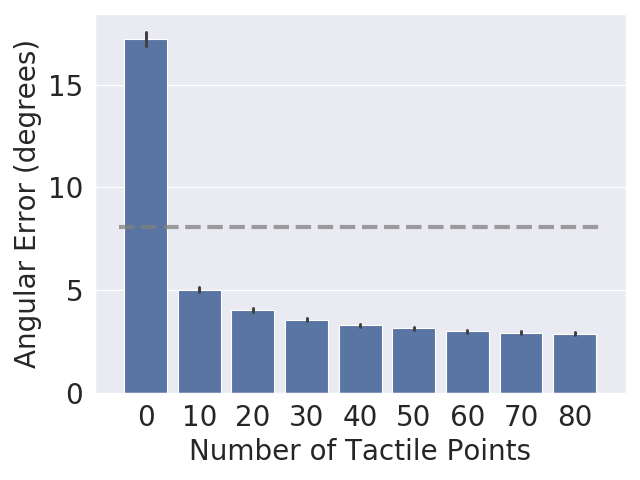}
    \end{subfigure}
    \caption{The performance of our approach under different levels of tactile contact points. The dashed gray line represents the performance of ViTa using 1000 tactile contact points.}
    \label{fig: tactile-points}
\end{figure}

The spatial resolution of real-world tactile sensors can vary significantly. Therefore, we analyze the performance of our model under different tactile contact points (Fig. \ref{fig: tactile-points}). %
It is noteworthy that our model, which was trained with 80 tactile points, demonstrates robust performance when presented with a reduced number of points. As expected, the performance of the model deteriorates as the number of tactile points is reduced and drops significantly when the tactile modality is removed entirely. It is worth noting that compared to ViTa, which requires tactile input, our model can still provide pose estimation even without tactile feedback, although with degraded performance. Upon analyzing the position error, we observe that, up to a reduction of 40 tactile points, our model outperforms ViTa, which uses 1000 points. The angular error results show that our model consistently outperforms ViTa.

\subsubsection{Ablation Studies}
\label{sec: ablations}
We perform ablation studies to examine the effectiveness of our design choices (Table \ref{tab: modalities}). %

\paragraph*{No-Shape-Completion} To evaluate the contribution of explicit shape completion, we remove the shape completion module from our proposed framework, which is achieved by feeding the partial latent vector $l_v^p$ to the pose estimator instead of the complete vector $\widehat{l_v^c}$. Our ablation result shows that by removing the shape completion module, the position error drops by 24.8\%, and the angular error drops by 30.0\%, suggesting, jointly and explicitly optimizing the shape and the pose during visuotactile pose estimation is effective.

\paragraph*{No-Vis-Gan} To analyze the gain from visual cues in the shape completion module, we remove the visual feature conditioning $f_i$. We observe that removing the visual conditioning from the shape completion module resulted in a significant deterioration of performance, highlighting the importance of incorporating visual cues. The study shows that, without visual cues, the partial geometry feature is ambiguous for inferring the complete shape under heavy hand occlusion.

\paragraph*{No-Vis-MLP} To examine the contribution of visual features in pose estimators, we remove the visual feature input from the pose estimators. %
We notice that removing the visual features degrades the angular error performance. This makes sense because our dataset contains symmetrical objects. The object geometry alone is insufficient for accurately determining the pose of symmetrical objects, such as a mustard bottle, which requires the utilization of visual features to distinguish between its front and back.

\paragraph*{No-Tactile \& No-Point-Cloud} Two separate studies were conducted to evaluate the contribution of the tactile points $P^\mathcal{T}$ (No-Tactile) and the point cloud from the vision sensor $P^\mathcal{D}$ (No-Point-Cloud). 
Our results suggest a significant drop in performance when either the tactile points or the point cloud input from the visual sensors is removed, emphasizing the significant contribution of both modalities to pose estimation.

\subsubsection{Real-world experiment}
We validate the sim-to-real robustness of our framework using our robot platform, with a subset of YCB objects that could be grasped by the Allegro Hand. The hand moves along a trajectory covering different poses. We apply a novel hand-grasping pose that doesn't exist in our training dataset, which shows our model's ability to generalize. Due to the excessive point cloud noise from the RGB-D sensor in the real world, we apply a point cloud statistical outlier filter as pre-processing using Open3D \cite{zhou_open3d_2018}. We consider 20 neighbors with a standard deviation ratio of 1. Our model efficiently operates at 11.2ms / 89Hz using an NVIDIA RTX 6000, thus capable of real-time deployment.

\begin{figure}[t!]
    \vspace*{0.15cm}
    \centering
    \begin{subfigure}{.32\columnwidth}
        \centering
        \includegraphics[width=\columnwidth, height=2.36cm]{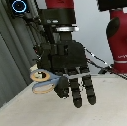}
    \end{subfigure}
    \begin{subfigure}{.32\columnwidth}
        \centering
        \includegraphics[width=\columnwidth, height=2.36cm]{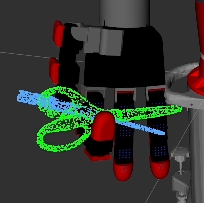}
    \end{subfigure}
    \begin{subfigure}{.32\columnwidth}
        \centering
        \includegraphics[width=\columnwidth, height=2.36cm]{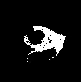}
    \end{subfigure}
    
    \begin{subfigure}{.32\columnwidth}
        \centering
        \includegraphics[width=\columnwidth, height=2.36cm]{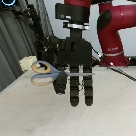}
    \end{subfigure}
    \begin{subfigure}{.32\columnwidth}
        \centering
        \includegraphics[width=\columnwidth, height=2.36cm]{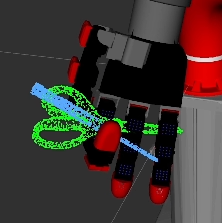}
    \end{subfigure}
    \begin{subfigure}{.32\columnwidth}
        \centering
        \includegraphics[width=\columnwidth, height=2.36cm]{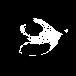}
    \end{subfigure}

    \begin{subfigure}{.32\columnwidth}
        \centering
        \includegraphics[width=\columnwidth, height=2.36cm]{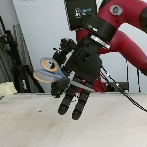}
        \caption{Observation}
    \end{subfigure}
    \begin{subfigure}{.32\columnwidth}
        \centering
        \includegraphics[width=\columnwidth, height=2.36cm]{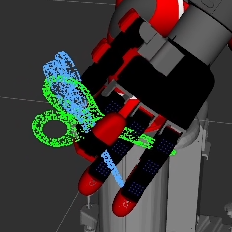}
        \caption{Estimation}
    \end{subfigure}
    \begin{subfigure}{.32\columnwidth}
        \centering
        \includegraphics[width=\columnwidth, height=2.36cm]{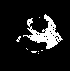}
        \caption{Segmentation}
    \end{subfigure}

    \caption{We compare our method (in green) against ViTa \cite{dikhale_visuotactile_2022} (in blue) in real-world. From left to right are the observation from the RGB-D sensor, the pose estimations, and the noisy partial segmentation task we obtain.}
    \label{fig: real-world}
\end{figure}

In Fig. \ref{fig: real-world}, we show three consecutive frames of the Allegro Hand holding a scissor. Our model could accurately estimate the 6D pose of the in-hand object, given a noisy and partial segmentation mask. In the bottom row, we demonstrate a failure case of our method where the estimated pose of the scissors is flipped 180 degrees due to the near-symmetrical geometry and visual feature of the scissors. We refer readers to our supplementary videos for our experiment videos that include more objects and real-time quantitative comparisons. 

\section{Conclusion} 
\label{sec:conclusion}
We presented ViHOPE, a novel framework for estimating the 6D pose of an in-hand object using visuotactile perception. We introduced explicit shape completion, which we hypothesize improves the accuracy of the 6D pose estimate by jointly optimizing both the shape and pose of the object. We validated our hypothesis by conducting experiments using a synthetic dataset of 11 YCB objects and compared ViHOPE's performance with state-of-the-art methods. Our results suggest a 35\% reduction in position error and a 64\% reduction in angular error in the pose estimation task compared to the state-of-the-art. We also demonstrated that ViHOPE outperforms state-of-the-art shape completion approaches by 265\% in terms of IoU and 88\% lower in CD. We presented the results of ablation studies that confirmed the contribution of explicit shape completion to the accuracy of the 6D pose estimate. To assess the practical viability of our framework in situations where access to high-resolution tactile sensors may be limited, we conducted experiments evaluating its performance under reduced tactile contact points. Our findings indicate that our framework outperforms the current state-of-the-art and can still produce reasonable pose estimates even in the absence of tactile feedback, although with a decreased performance. Finally, we validated the sim-to-real robustness of ViHOPE in a real-world robot experiment, suggesting its success in estimating the 6D pose of an object in real-world settings.

In our study, we used an empirically determined voxel resolution of $32^3$. Using a coarser grid would compromise performance, while a finer grid would introduce computational overhead and latency. The voxel resolution is crucial in capturing the level of detail in the object's shape. Higher resolutions capture finer geometric features. This aspect becomes particularly important for objects whose accurate pose determination depends on distinguishing geometric features smaller than the selected voxel grid resolution. Moving forward, we plan to explore octree representations~\cite{tatarchenko_octree_2017, li2023stereovoxelnet} that enable higher resolutions while maintaining computational efficiency.

We focus on instance-level pose estimation in this letter. In the future, we are interested in extending our pose estimator to work on more challenging scenarios, for example, lack of annotated data \cite{zhang_self-supervised_2022}, and category-level pose estimation \cite{wang_normalized_2019}. Another interesting future direction is incorporating other sensory information, such as pressure, into our framework. Another direction that can be pursued is the use of methods such as long short-term memory networks to use temporal coherence to filter out erroneous estimates.

\footnotesize{
\bibliographystyle{IEEEtranN}
\bibliography{custom}
}

\end{document}